\documentclass{aamas2016}

\usepackage{url}
\usepackage{amsfonts}
\usepackage{graphicx}
\usepackage{multirow}
\usepackage{sistyle}
\usepackage{color}
\usepackage[normalem]{ulem}

\SIthousandsep{,}

\begin{document}


\title{State of the Art Control of Atari Games\\ Using Shallow Reinforcement Learning}



%
%
%
%

%

\numberofauthors{4}

\author{
Yitao Liang$^\dagger$, Marlos C. Machado${^\ddagger}$, Erik Talvitie$^\dagger$, and Michael Bowling${^\ddagger}$\\
  \begin{tabular}[t]{@{}c@{}}
  \affaddr{$^\dagger$Franklin \& Marshall College}\\
  \affaddr{Lancaster, PA, USA}\\
  \email{\{yliang, erik.talvitie\}@fandm.edu}
  \end{tabular}\nobreak\qquad
  \begin{tabular}[t]{@{}c@{}}
  \affaddr{$^\ddagger$University of Alberta}\\
  \affaddr{Edmonton, AB, Canada}\\
  \email{\{machado, mbowling\}@ualberta.ca}
  \end{tabular}
}


\maketitle

\begin{abstract}
The recently introduced Deep Q-Networks (DQN) algorithm has gained attention as one of the first successful combinations of deep neural networks and reinforcement learning. Its promise was demonstrated in the Arcade Learning Environment (ALE), a challenging framework composed of dozens of Atari 2600 games used to evaluate general competency in AI. It achieved dramatically better results than earlier approaches, showing that its ability to learn good representations is quite robust and general. This paper attempts to understand the principles that underlie DQN's impressive performance and to better contextualize its success. We systematically evaluate the importance of key representational biases encoded by DQN's network by proposing simple linear representations that make use of these concepts. Incorporating these characteristics, we obtain a computationally practical feature set that achieves competitive performance to DQN in the ALE. Besides offering insight into the strengths and weaknesses of DQN, we provide a generic representation for the ALE, significantly reducing the burden of learning a representation for each game. Moreover, we also provide a simple, reproducible benchmark for the sake of comparison to future work in the~ALE.
\end{abstract}




\section{Introduction}
In the reinforcement learning (RL) problem an agent autonomously learns a behavior policy from experience in order to maximize a provided reward signal. Most successful RL approaches have relied upon the engineering of problem-specific state representations, diminishing the agent as fully autonomous and reducing its flexibility. The recent Deep Q-Network (DQN) algorithm~\cite{Mnih15} aims to tackle this problem, presenting one of the first successful combinations of RL and deep convolutional neural-networks (CNN) \cite{Krizhevsky12,LeCun98}, which are proving to be a powerful approach to representation learning in many areas. DQN is based upon the well-known Q-learning algorithm \cite{Watkins92} and uses a CNN to simultaneously learn a problem-specific representation and estimate a value function.

\begin{figure}[t]
  \centering
  \includegraphics[width=0.45\textwidth]{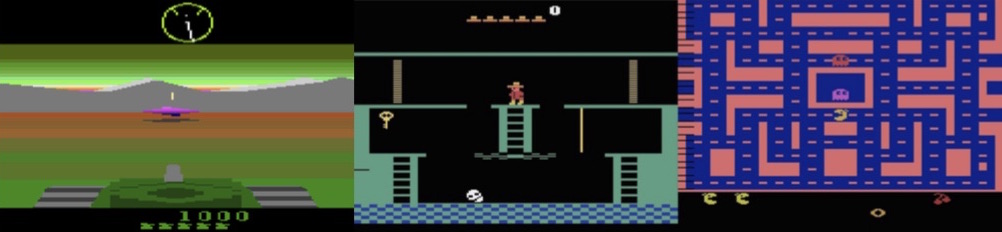}
  \caption{Examples from the ALE (left to right: \textsc{Battle Zone, Montezuma's Revenge, Ice Hockey}).}
  \label{fig:ale}
\end{figure}

Games have always been an important testbed for AI, frequently being used to demonstrate major contributions to the field \cite{Bowling15,Campbell02,Ferrucci10,Schaeffer07,Tesauro95}. DQN follows this tradition, demonstrating its success by achieving human-level performance in the majority of games within the Arcade Learning Environment (ALE)~\cite{Bellemare13}. The ALE is a platform composed of dozens of qualitatively diverse Atari 2600 games. As pictured in Figure~\ref{fig:ale}, the games in this suite include first-person perspective shooting games (e.g. \textsc{Battle Zone}), platforming puzzle games (e.g. \textsc{Montezuma's Revenge}), sports games (e.g. \textsc{Ice Hockey}), and many other genres. Because of this diversity, successful approaches in the ALE necessarily exhibit a degree of robustness and generality. Further, because it is based on problems designed by humans for humans, the ALE inherently encodes some of the biases that allow humans to successfully navigate the world. This makes it a potential stepping-stone to other, more complex decision-making problems, especially those with visual input.

DQN's success in the ALE rightfully attracted a great deal of attention. For the first time an artificial agent achieved performance comparable to a human player in this challenging domain, achieving by far the best results at the time and demonstrating generality across many diverse problems. It also demonstrated a successful large scale integration of deep neural networks and RL, surely an important step toward more flexible agents. That said, problematic aspects of DQN's evaluation make it difficult to fully interpret the results. As will be discussed in more detail in Section \ref{comparison_dqn}, the DQN experiments exploit non-standard game-specific prior information and also report only one independent trial per game, making it difficult to reproduce these results or to make principled comparisons to other methods.Independent re-evaluations have already reported different results due to the variance in the single trial performance of DQN (e.g. \cite{Hausknecht15}). Furthermore, the comparisons that Mnih et al. did present were to benchmark results using far less training data. Without an adequate baseline for what is achievable using simpler techniques, it is difficult to evaluate the cost-benefit ratios of more complex methods like DQN.

In addition to these methodological concerns, the evaluation of a complicated method such as DQN often leaves open the question of which of its properties were most important to its success. While it is tempting to assume that the neural network must be discovering insightful tailored representations for each game, there is also a considerable amount of domain knowledge embedded in the very structure of the network.  We can identify three key structural biases in DQN's representation. First, CNNs provide a form of spatial invariance not exploited in earlier work using linear function approximation. DQN also made use of multiple frames of input allowing for the representation of short-range non-Markovian value functions. Finally, small-sized convolutions are quite well suited to detecting small patterns of pixels that commonly represent objects in early video game graphics. These biases were not fully explored in earlier work in the ALE, so a natural question is whether non-linear deep network representations are key to strong performance in ALE or whether the general principles implicit in DQN's network architecture might be captured more simply.

The primary goal of this paper is to systematically investigate the sources of DQN's success in the ALE. Obtaining a deeper understanding of the core principles influencing DQN's performance and placing its impressive results in perspective should aid practitioners aiming to apply or extend it (whether in the ALE or in other domains). It should also reveal representational issues that are key to success in the ALE itself, potentially inspiring new directions of research. 

We perform our investigation by elaborating upon a simple linear representation that was one of DQN's main comparison points, progressively incorporating the representational biases identified above and evaluating the impact of each one. This process ultimately yields a fixed, generic feature representation able to obtain performance competitive to DQN in the ALE, suggesting that the general form of the representations learned by DQN may in many cases be more important than the specific features it learns. Further, this feature set offers a simple and computationally practical alternative to DQN as a platform for future research, especially when the focus is not on representation learning. Finally, we are able provide an alternative benchmark that is more methodologically sound, easing reproducibility and comparison to future work.

\section{Background}
In this section we introduce the reinforcement learning problem setting and describe existing results in the ALE.

\subsection{Reinforcement Learning }
In the {\em reinforcement learning} (RL) problem~\cite{Sutton98,Szepesvari10} an agent interacts with an
unknown environment and attempts to maximize a ``reward'' signal. The
environment is commonly formalized as a {\em Markov decision process} (MDP) $\mathcal{M}$ defined as a 5-tuple $\mathcal{M} = \langle \mathcal{S}, \mathcal{A}, \mathcal{R}, P, \gamma\rangle$. At time $t$ the agent is in the state $s_t \in \mathcal{S}$ where it takes an action $a_t \in \mathcal{A}$ that leads to the next state $s_{t+1} \in \mathcal{S}$ according to the transition probability kernel $P$, which encodes $Pr(s_{t+1} | s_t, a_t)$. The agent also observes a reward $r_{t+1} \sim \mathcal{R}(s_t, a_t, s_{t+1})$. The agent's goal is to learn the {\em optimal policy}, a conditional distribution $\pi(a | s)$ that maximizes the {\em state value function} $V^\pi(s) \doteq \mathbb{E}_\pi \big[\sum_{k=0}^{\infty} \gamma^k r_{t+k+1} | s_t = s\big]$ for all $s \in \mathcal{S}$, where $\gamma \in [0, 1)$ is known as the {\em discount factor}.

As a critical step toward improving a given policy $\pi$, it is common for reinforcement learning algorithms to learn a {\em state-action value function}, denoted: 
$$Q^\pi (s,a) \doteq \mathbb{E}_\pi \big[\mathcal{R}(s_t, a_t, s_{t+1})~+~\gamma V^\pi(s_{t+1})~\big|s_t\!=\!s,a_t\!=\!a\big].$$ 
However, in large problems it may be infeasible to learn a value for each state-action pair. To tackle this issue agents often learn an approximate value function: $Q^\pi(s, a; \theta) \approx Q^\pi(s, a)$. A common approach uses linear function approximation (LFA) where $Q^\pi(s, a; \theta) = \theta^\top \phi(s,a)$, in which $\theta$ denotes the vector of weights and $\phi(s,a)$ denotes a static feature representation of the state $s$ when taking action $a$. However, this can also be done through non-linear function approximation methods, including neural networks.

One of the most popular reinforcement learning algorithms is Sarsa($\lambda$) \cite{Rummery94}. It consists of learning an approximate action-value function while following a continually improving policy $\pi$. As the states are visited, and rewards are observed, the action-value function is updated and consequently the policy is improved since each update improves the estimative of the agent's expected return from state $s$, taking action $a$, and then following policy $\pi$ afterwards, \emph{i.e.} $Q^\pi(s, a)$. The Sarsa($\lambda$) update equations, when using function approximation, are:
\begin{eqnarray*}
  \delta_t &=& r_{t+1} + \gamma Q(s_{t+1}, a_{t+1}; \vec{\theta}_t) - Q(s_t, a_t; \vec{\theta}_t)\\
  \vec{e_t} &=& \gamma \lambda \vec{e}_{t-1} + \nabla Q(s_t, a_t; \vec{\theta}_t)\\
  \vec{\theta}_{t+1} &=& \vec{\theta}_t + \alpha \delta_t \vec{e}_t,
\end{eqnarray*}
where $\alpha$ denotes the step-size, the elements of $\vec{e}_t$ are known as the {\em eligibility traces}, and $\delta_t$ is the {\em temporal difference error}. Theoretical results suggest that, as an on-policy method, Sarsa($\lambda$) may be more stable in the linear function approximation case than off-policy methods such as Q-learning, which are known to risk divergence \cite{Baird95,Gordon00,Melo08}. These theoretical insights are confirmed in practice; Sarsa($\lambda$) seems to be far less likely to diverge in the ALE than Q-learning and other off-policy methods \cite{Defazio13}.

\subsection{Arcade Learning Environment}

In the Arcade Learning Environment agents have access only to sensory information (160 pixels wide by 210 pixels high images) and aim to maximize the score of the game being played using the 18 actions on a standard joystick without game-specific prior information. Note that in most games a single screen does not constitute Markovian state. That said, Atari 2600 games are deterministic, so the entire history of interaction fully determines the next screen. It is most common for ALE agents to base their decisions on only the most recent few screens.

\subsubsection{Linear Value Function Approximation}
Early work in the ALE focused on developing new generic feature representations to be used with linear RL methods. Along with introducing the ALE itself, Bellemare et al.~\cite{Bellemare13} presented an RL benchmark using four different feature sets obtained from the game screen. \emph{Basic} features tile the screen and check if each of the available colors in the Atari 2600 are active in each tile. \emph{BASS} features add pairwise combinations of Basic features. \emph{DISCO} features attempt to detect and classify objects on the screen in order to infer their positions and velocities. \emph{LSH} simply applies Locally Sensitive Hashing \cite{Gionis99} to raw Atari 2600 screens. Bellemare, Veness, and Bowling proposed an extension to the Basic feature set which involved identifying which parts of the screen were under the agents' direct control. This additional information is known as \emph{contingency awareness} \cite{Bellemare12a}. 

These feature sets were, for some time, the standard representations for Atari 2600 games, being directly used in other work \cite{Defazio13,Lipovetzky15,Machado15} or serving as the base for more elaborate feature sets \cite{Bellemare12b}. The feature representations presented in this paper follow the spirit of this early work. We use Basic features (formally described below) as a starting point, and attempt to capture pairwise spatial relationships between objects on the screen.\\

\noindent {\bf Basic Features:} To obtain Basic features we first divide the Atari 2600 screen into $16\times14$ tiles of size $10 \times 15$ pixels. For every tile $(c, r)$ and color $k$, where $c \in \{1, \ldots, 16\}$, $r \in \{1, \ldots, 14\}$, and $k \in \{1, \ldots, 128\}$, we check whether color $k$ is present within the tile $(c, r)$, generating the binary feature $\phi_{c,r,k}$. Intuitively, Basic features encode information like ``there is a red pixel within this region''. There are $16\times14\times128 = \num{28672}$ Basic features in total.

It can be computationally demanding to work directly with $160 \times 210$ pixel screens with a 128 color palette. To make the screen more sparse, following previous work (\emph{e.g.} \cite{Bellemare13,Bellemare12a}) we subtract the background from the screen at each frame before processing it. The background is precomputed using \num{18000} samples obtained from random trajectories.

\subsubsection{Non-Linear Value Function Approximation}

The use of non-linear function approximation in the ALE is more recent. Veness et al.~\cite{Veness15}, for example, propose the use of compression techniques as a policy evaluation approach in RL, evaluating their method in the ALE. Soon after, the research landscape changed due to the success of using deep learning techniques to approximate the action-value functions, demonstrated by the DQN algorithm \cite{Mnih13,Mnih15}. DQN achieved  75\% of a human player's score in the majority of games and this inspired much more work in the ALE using deep learning (\emph{e.g.} \cite{Guo14,Schulman15,Sprague15}). 

DQN employs a deep convolutional neural network to represent the state-action value function $Q$. A deep convolutional network uses weight sharing to make it practical to learn a function of high-dimensional image input. Specifically, when processing the image a small network focused on a small region of the screen called a {\em filter} is applied at multiple positions on the screen. Its output at each position forms a new (smaller) image, which can then be processed by another filter, and so on. Multiple layers of convolution allow the network to detect patterns and relationships at progressively higher levels of abstraction. DQN's network uses three layers of convolution with multiple filters at each layer. The final result of those convolutions is then processed via a more standard fully connected feed-forward network. The convolutional aspect of the network allows the network to detect relationships like ``A bullet is near an alien here,'' where ``bullet'' and ``alien'' can be position invariant concepts. The non-linear global layers allow it to represent whole-screen concepts such as ``A bullet is near an alien somewhere.'' This is the inspiration of our spatially invariant features (Section~3). Further, Mnih et al. actually provide the four most recent images as input to their network, allowing it to detect relationships through time (such as movement). This motivates our short-order-Markov features (Section~4). Finally, note that Basic features are essentially a simple convolution where the filters simply detect the presence of each color in their range. DQN's more sophisticated filters motivate our improvement of object detection for base features (Section~5).

\section{Spatial Invariance}
Recall that Basic features detect the presence of each color in various regions of the screen. As discussed above, this can be seen as analogous to a single convolution with a crude filter. The BASS feature set, which was amongst the best performing representations before DQN's introduction encodes pairwise relationships between Basic features, still anchored at specific positions, so it is somewhat analogous to a second convolutional layer. But in many games the absolute positions of objects are not as important as their relative positions to each other. To investigate the effect of ignoring absolute position we impose a non-linearity over BASS (analogous to DQN's fully connected layers), specifically taking the max of BASS features over absolute position.

We call the resulting feature set Basic Pairwise Relative Offsets in Space (B-PROS) because it captures pairwise {\em relative} distances between objects in a single screen. More specifically, a B-PROS feature checks if there is a pair of Basic features with colors $k_1,k_2 \in \{1, \ldots, 128\}$ separated by an offset $(i, j)$, where $-13 \leq i \leq 13$ and $-15 \leq j \leq 15$. If so, $\phi_{k_1,k_2,(i,j)}$ is set to 1, meaning that a pixel of color $k_1$ is contained within some block $(c,r)$ and a pixel of color $k_2$ is contained within the block $(c+i, r+j)$. Intuitively, B-PROS features encode information like ``there is a yellow pixel three tiles below a red pixel''. The computational complexity of generating B-PROS features is similar to that of BASS, though ultimately fewer features are generated. Note that, as described, B-PROS contains redundant features (e.g. $\phi_{1,2,(4,0)}$ and $\phi_{2,1,(-4,0)}$), but it is straightforward to eliminate them. The complete feature set is composed of the Basic features and the B-PROS features. After redundancy is eliminated, the B-PROS feature set has $\num{6885440}$ features in total ($\num{28672} + ((31 \times 27 \times 128^2-128)/2+128)$). 

Note that there have been other attempts to represent pairwise spatial relationships between objects, for instance DISCO \cite{Bellemare13} and contingency awareness features \cite{Bellemare12a}. However, these existing attempts are complicated to implement, demanding to compute, and less effective (as will be seen), most likely due to unreliable estimates of object positions.

\subsection{Empirical Evaluation}\label{empirical_lfa}

Our first set of experiments compares B-PROS to the previous state of the art in linear representations for the Atari 2600 \cite{Bellemare13,Bellemare12a} in order to evaluate the impact of the non-linearity applied to BASS. For the sake of comparison we follow Bellemare et al.'s methodology~\cite{Bellemare13}. Specifically, in 24 independent trials the agent was trained for \num{5000} episodes. After the learning phase we froze the weights and evaluated the learned policy by recording its average performance over 499 episodes. We report the average evaluation score across the 24 trials. Following Bellemare et al., we defined a maximum length for each episode: \num{18000} frames, \emph{i.e.} five minutes of real-time play. Also, we used a frame-skipping technique, in which the agent selects actions and updates its value function every $x$ frames, repeating the selected action for the next $x-1$ frames. This allows the agent to play approximately $x$ times faster. Following Bellemare et al. we use $x = 5$ (DQN also uses a similar procedure).

We used Sarsa($\lambda$) with replacing traces and an $\epsilon$-greedy policy. We performed a parameter sweep over nine games, which we call ``training'' games. The reported results use a decay rate $\gamma=0.99$, an exploration rate $\epsilon=0.01$, a step-size $\alpha = 0.5$ and eligibility decay rate $\lambda=0.9$.

\begin{table}[t]
\begin{center}
  \caption{Comparison of linear representations. Bold denotes the largest value between B-PROS and Best Linear [2]. See text for more details.}
  \label{table:short_results_lfa}
  \tiny{
  \begin{tabular}{|r|r|r|rl|}
    \hline  
		Game                     &Best Linear    &CAF       &\multicolumn{2}{c|}{B-PROS (std. dev.)}\\ \hline \hline
		\textsc{Asterix}         & 987.3         &1332.0    & {\bf 4078.7}   & (1016.0)     \\ \hline
		\textsc{Beam Rider}      & 929.4         &1742.7    & {\bf 1528.7}   & (504.4)      \\ \hline
		\textsc{Breakout}        & 5.2           &6.1       & {\bf 13.5}     & (11.0)       \\ \hline
		\textsc{Enduro}          & 129.1         &159.4     & {\bf 240.5}    & (24.2)       \\ \hline
		\textsc{Freeway}         & 16.4          &19.97     & {\bf 31.0}     & (0.7)        \\ \hline
		\textsc{Pong}            & -19.2         &-17.4     & {\bf 4.9 }     & (6.6)        \\ \hline
		\textsc{Q*Bert}          & 613.5         &960.3     & {\bf 1099.8}   & (740.0)      \\ \hline
		\textsc{Seaquest}        & 664.8         &722.89    & {\bf 1048.5}   & (321.2)      \\ \hline
		\textsc{Space Invaders}  & 250.1         &267.93    & {\bf 384.7}    & (87.2)       \\ \hline
   \end{tabular}
  }
  \end{center}
\end{table}

Table~\ref{table:short_results_lfa} compares our agent to existing linear agents in our set of training games (note that \textsc{Breakout}, \textsc{Enduro}, \textsc{Pong}, and \textsc{Q*Bert} were not training games for the earlier methods). ``Best Linear'' denotes the best performance obtained among four different feature sets: Basic, BASS, DISCO and LSH \cite{Bellemare13}. For reference, in the ``CAF'' column we also include results obtained by contingency awareness features \cite{Bellemare12a}, as reported by Mnih et al.~\cite{Mnih15}. Note, that these results are not directly comparable because their agent was given \num{10000} episodes of training (rather than \num{5000}).  

B-PROS' performance surpasses the original benchmarks by a large margin and, except in one game, even surpasses the performance of CAF, despite being far simpler and training with half as much data. Further, some of the improvements represent qualitative breakthroughs. For instance, in \textsc{Pong}, B-PROS allows the agent to win the game with a score of 20-15 on average, while previous methods rarely scored more than a few points. In \textsc{Enduro}, the earlier methods seem to be stymied by a sudden change in dynamics after approximately 120 points, while B-PROS is consistently able to continue beyond that point. 

When comparing these features sets for all 53 games B-PROS performs better on average than all of Basic, BASS, DISCO, and LSH in 77\% (41/53) of the games. Even with less training data, B-PROS performs better on average than CAF in 77\% (38/49) of the games in which CAF results were reported\footnote{For reference, the full results with 53 games are available in Table~3 in the Appendix.}. The dramatic improvement yielded by B-PROS clearly indicates that focusing on relative spatial relationships rather than absolute positions is vital to success in the ALE (and likely in other visual domains as well).

\section{Non-Markovian Features} \label{bprost}
B-PROS is capable of encoding relative distances between objects but it fails to encode movement, which can be a very important aspect of Atari games. For instance, the agent may need to know whether the ball is moving toward or away from the paddle in \textsc{Pong}. Previous linear representations similarly relied mainly on the most recent screen. In contrast, Mnih et al. use the four most recent screens as input, allowing DQN to represent short-order-Markov features of the game screens. In this section we present an extension to B-PROS that takes a similar approach, extracting information from the two most recent screens.

Basic Pairwise Relative Offsets in Time (B-PROT) features represent pairwise relative offsets between Basic features obtained from the screen five frames in the past and Basic features from the current screen ({\em i.e.} PROT features). More specifically, for every pair of colors  $k_1, k_2 \in \{1,\ldots,128\}$ and every offset $(i, j)$, where $-13 \leq i \leq 13$ and $-15 \leq j \leq 15$, a binary B-PROT feature $\phi_{k_1,k_2,(i,j)}$ is 1 if a pixel of color $k_1$ is contained within some block $(c+i, r+j)$ on the screen five frames ago and a pixel of color $k_2$ is contained within the block $(c, r)$ in the current screen.

The B-PROST feature set contains Basic, B-PROS, and B-PROT features. Note that there are roughly twice as many B-PROT features as B-PROS because there are no redundant offsets. As such, B-PROST has a total of \num{20598848} sparse, binary features ($\num{6885440} + 31 \times 27 \times 128^2$).

\subsection{Empirical Evaluation}
B-PROS outperformed all the other linear architectures in the previous experiment. Subsequent extensions will be primarily compared to DQN (see Section ~\ref{comparison_dqn}). As such, in these experiments, we adopted an evaluation protocol similar to Mnih et al.'s. Each agent was trained for \num{200000000} frames (equivalent to \num{40000000} decisions) over 24 independent trials. The learned policy in each trial was evaluated by recording its average performance in 499  episodes with no learning. We report the average evaluation score over the 24 trials. In an effort to make our results comparable to DQN's we also started each episode with a random number of ``no-op'' actions and restricted the agent to the minimal set of actions that have a unique effect in each game.

The first two columns of Table \ref{table:short_results_bpro} present results using B-PROS and B-PROST in the training games. B-PROST outperforms B-PROS in all but one of the training games. One particularly dramatic improvement occurs in \textsc{Pong}; B-PROS wins with a score of 20-9, on average, while B-PROST rarely allows the opponent to score at all. Another result worth noting is in \textsc{Enduro}. The randomized initial conditions seem to have significantly harmed the performance of B-PROS in this game, but B-PROST seems to be robust to this effect. When evaluated over all 49 games evaluated by Mnih et al.  the average score using B-PROST is higher than that using B-PROS in 82\% of the games (40/49)\footnote{The full results are available in Table~4 in the Appendix.}. This clearly indicates the critical importance of non-Markov features to success in the ALE.

Before making a final comparison with DQN, we will make one more improvement to our representation.

\begin{table}[t]
\begin{center}
  \caption{Comparison of relative offset features. Bold indicates the best average of the three columns. The $^\dagger$ indicates significant differences between B-PROST and Blob-PROST.}
  \label{table:short_results_bpro}
\tiny{
\begin{tabular}{|r|l|l|l|}
\hline
\multirow{2}{*}{Game}         &\multicolumn{1}{c|}{{B-PROS}}             &\multicolumn{1}{c|}{B-PROST}                 &\multicolumn{1}{c|}{Blob-PROST} \\
                              &\multicolumn{1}{c|}{{Avg. (std. dev.)}}   &\multicolumn{1}{c|}{{Avg. (std. dev.)}}      &\multicolumn{1}{c|}{{Avg. (std. dev.)}} \\ \hline \hline

\textsc{Asterix}              &8194.3  (1802.9)    &{\bf 8928.8}$^\dagger$ (1714.5)  &4400.8 (1201.3)  \\ \hline
\textsc{Beam Rider}           &1686.9 (255.2)     &1808.9 (309.2)   &{\bf 1902.3} (423.9)   \\ \hline
\textsc{Breakout}             &7.1 (1.7)       &15.0 (4.7)     &{\bf 46.7}$^\dagger$ (45.7)    \\ \hline
\textsc{Enduro}               &34.9 (71.5)      &207.7 (23.1)    &{\bf 257.0}$^\dagger$ (79.8)    \\ \hline
\textsc{Freeway}              &23.8 (6.7)       &29.1 (6.3)     &{\bf 31.5}$^\dagger$ (1.2)     \\ \hline
\textsc{Pong}                 &10.9 (5.2)       &18.9 (1.3)     &{\bf 20.1}$^\dagger$ (0.5)     \\ \hline
\textsc{Q*Bert}               &3647.8 (1273.3)    &3608.7 (1129.0)  &{\bf 6946.8}$^\dagger$ (3036.4)  \\ \hline
\textsc{Seaquest}             &1366.1 (445.9)     &1636.5 (519.5)   &{\bf 1664.2} (440.4)   \\ \hline
\textsc{Space Invaders}       &505.1 (130.4)     &582.9 (139.0)   &{\bf 723.1}$^\dagger$ (101.5)   \\ \hline
\end{tabular}
}
\end{center}
\end{table}

\section{Object Detection}
Because Basic features encode the positions of individual pixels on the screen, both B-PROS and B-PROST features struggle to distinguish which pixels are part of the same object. DQN's network is capable of learning far more subtle filters. In order to measure the impact of improved low-level object detection, we consider a simple extension to Basic that exploits the fact that Atari screens often contain several contiguous regions of pixels of the same color. We call such regions ``blobs". Rather than directly represent coarsened positions of pixels, we first process the screen to find a list of blobs. Blob features then represent coarsened positions of blobs on the screen. Changing the ``primitive'' features from Basic to Blob yields the Blob-PROST feature set.

Note that blobs are a simplification; in many Atari games, as would be true in more natural images, objects consist of multiple close but separate blobs. Grouping only strictly contiguous pixels into each blob may generate redundant blobs that all represent a single object. As a simple means to address this we add a tolerance to the contiguity condition, \emph{i.e.} we consider pixels that are in the same $s\times s$ pixel square to be contiguous. This approach has an inherent trade-off. On the one hand, with sufficiently large $s$ we may successfully represent each object with few blobs. Also, by reducing the number of blobs, we substantially decrease the number of primitive features, making Blob-PROS and Blob-PROT features easier to compute. On the other hand, if $s$ is too high then multiple distinct objects may be grouped together. In our experiments we set $s$ to be 6 after an informal search using the set of training games. It is very likely that with a more systematic selection of $s$ one can obtain better results than those reported here.

We define the position of a blob as the centroid of the blob's smallest bounding box. To generate the features, we first generate Blob features that are analogous to Basic features: we divide the screen into tiles of $4\times7$ pixels and then for every color $k$ and block $(c,r)$, where $c \in \{1, \ldots, 40\}, r \in \{1, \ldots, 30\}$ and $k \in \{1,\ldots, 128\}$, the Blob feature $\phi_{c, r, k}$ is 1 if the block $(c, r)$ contains the centroid for some blob of color $k$. Note that we use a finer resolution than Basic. This is feasible only because of the extreme sparsity of blobs on the screen. The number of blobs in one frame ranged from \num{6} (\textsc{Pong}) to \num{412} (\textsc{Battle Zone}). This sparsity also makes the background subtraction step unnecessary; the background typically reduces to a handful of blobs itself. 

The Blob-PROST feature set is then constructed from these primitive features in the same manner as B-PROST. There are \num{153600} Blob features ($40 \times 30 \times 128$), \num{38182976} Blob-PROS features ($(79\times59\times128^2-128)/2+128$), and \num{76365824} Blob-PROT features ($79\times59\times128^2$). This yields a total of \num{114702400} Blob-PROST features. Though the number of possible features is very large, most would never be generated due to the sparsity of blobs.

\subsection{Empirical Evaluation}

The last column of Table \ref{table:short_results_bpro} presents results using Blob-PROST in the training games. In all but one of the training games, Blob-PROST outperformed both B-PROS and B-PROST on average\footnote{The implementation of the three feature sets we introduced, as well as the code used in our experiments is available at \url{https://github.com/mcmachado/b-pro}.}. In 6 of the 9, Blob-PROST's average performance was statistically significantly better than B-PROST (using Welch's t-test with $p < 0.05$). \textsc{Asterix} is a game in which Blob-PROST notably fails; perhaps blob detection lumps together distinct objects in a harmful way. \textsc{Q*Bert} is a notable success; the scores of B-PROS and B-PROST indicate that they rarely complete the first level, while Blob-PROST's score indicates that it consistently clears the first level. Out of the 49 games evaluated by Mnih et al., Blob-PROST's average performance was higher than that of both B-PROS and B-PROST in 59\% (29/49) of the games. Its performance was statistically significantly higher than that of B-PROST in 47\% (23/49) of the games, showing the clear contribution of even simple improvements in object detection. Though object detection itself is not a feature, it make all features based upon it carry more meaningful information that helps our agents better ``interpret'' the screen.

\section{Comparison with DQN} \label{comparison_dqn}
The next set of experiments compare Blob-PROST to DQN, showing that the three enhancements investigated above largely explain the performance gap between DQN and earlier LFA approaches. In performing this comparison, we necessarily confront problematic aspects of the original DQN evaluation (some of which we adopt for comparison's sake). In Section~\ref{benchmark} we present a more methodologically sound ALE benchmark for comparison to future work.

\subsection{DQN Evaluation Methodology} 

In each game, Mnih et al. trained a DQN agent once for \num{200000000} frames. Rather than evaluating the learned policy after the whole learning phase completed (as in earlier work), they evaluated their agent's performance every \num{4000000} frames and selected the best performing weights. They reported the average and standard deviation of the scores of this best learned policy in $30$ evaluation episodes. 

Mnih et al. also restricted their agent to the minimal set of actions that have a unique effect in each game, game-specific prior information not exploited by earlier work in the ALE. Since in most games the minimal action set contains fewer actions than the full action set, agents with access to the minimal set may benefit from a faster effective learning rate (this is discussed further in Section~\ref{benchmark}).

To avoid overfitting to the Atari's determinism, DQN added a  random number of ``no-op" actions to the beginning of each training and evaluation episode: the number of ``no-ops'' was uniformly randomly selected from $\{1, \ldots, 30\}$, before a new episode officially began.

Mnih et al. also modified when episodes ended. In many Atari games the player has a number of ``lives'' and the game ends when they are exhausted. DQN employed the same episode termination criteria as Bellemare et al.~\cite{Bellemare13}, {\em i.e.} the end of the game or expiration of the 5 minute time limit, but during training also terminated episodes when the agent lost a life (another form of game-specific prior information). We have not specifically evaluated the impact of this mechanism, but one might speculate that in games that have ``lives'' (e.g. \textsc{Breakout}, \textsc{Space Invaders}) this modification could more easily generate agents with an ``avoid death'' policy.

\begin{figure}
\begin{center}
  \includegraphics[scale=0.15]{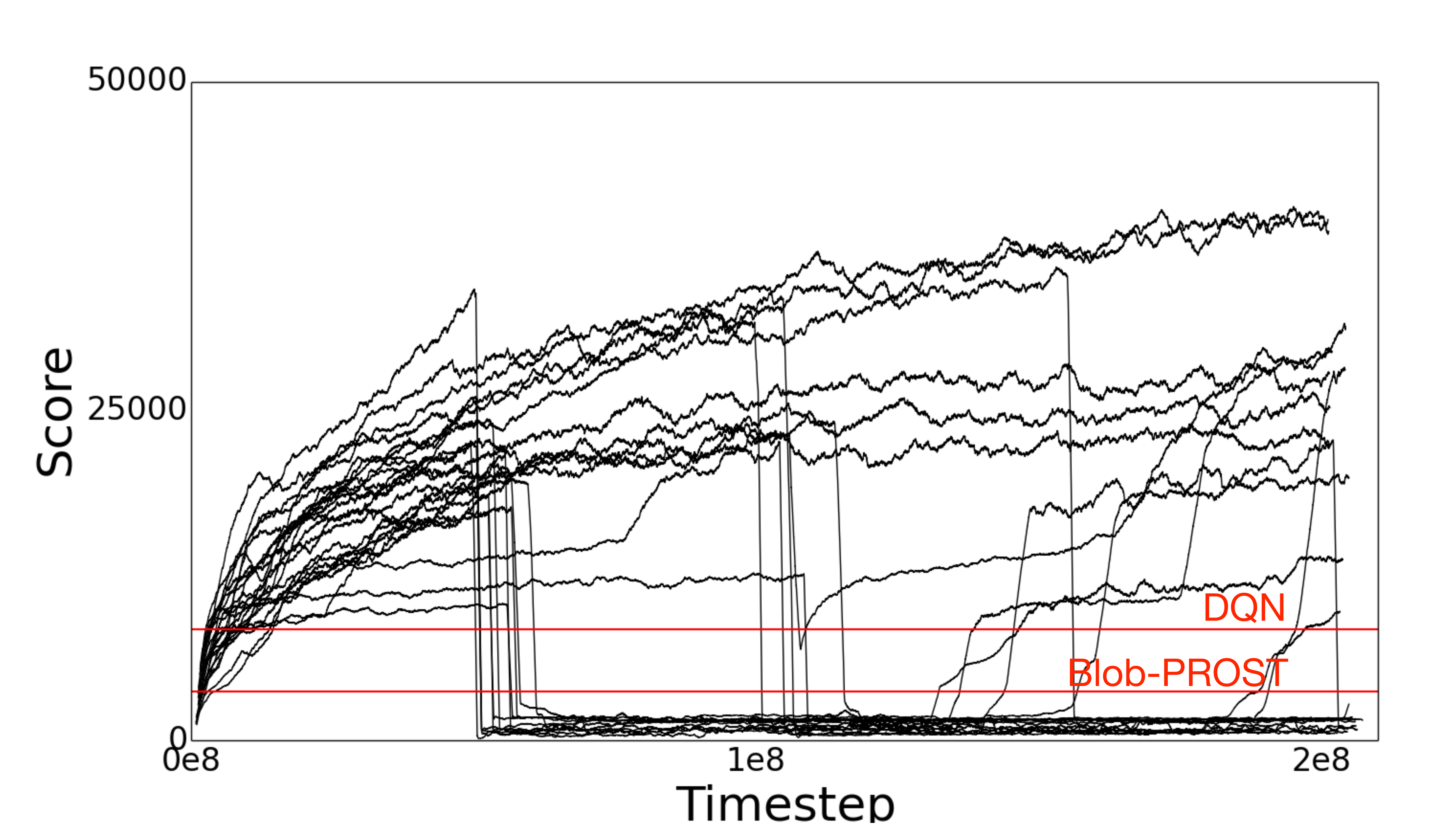}
  \caption{Learning curves for Blob-PROST on the game \textsc{Up 'n Down}, smoothed over 500 episodes.}
  \label{fig:upndown}
\end{center}
\end{figure}

The DQN experimental methodology possesses three main problematic flaws. First, one cannot make statistical comparisons from a single trial.  While Mnih et al. report a standard deviation, this only represents the amount of variation observed when executing the one chosen policy, and not variation observed over different independent executions of the inherently stochastic algorithm.  DQN's consistently high performance across games suggests it does indeed regularly perform well, but the reported performance numbers on any particular game may not be representative of their expected performance.  Second, selecting the best performing parameterization over the whole learning period is a significant deviation from typical RL methodology.  Such a choice may be masking instability issues where learning performance is initially strong but later goes awry.  This effect can occur even using comparatively stable linear function approximators.  For example, Figure~\ref{fig:upndown} shows 24 independent learning trials from the game {\sc Up 'n Down} using our linear Blob-PROST agent.  Many learning trials in this game exhibit an agent steadily improving its performance before a sudden plummet from which it does not recover. The reported performance of DQN and Blob-PROST is also depicted, showing the impact of such instability: most trials perform much better than the reported average at some point. We contend that general competency should include consistent and stable improvement over time, best measured by an agent's performance at the end of training. Third, using designer-provided game-specific prior knowledge in the form of the minimal action set and termination on ``death''  is counter to the goal of evaluating general competency. Emerging follow-up work~\cite{Mnih16} has begun to address some of these issues by using the final network weights for evaluation and by reporting the average score over five independent trials. 

\subsection{Comparing Blob-PROST and DQN}
As discussed in Section~\ref{bprost}, for comparison's sake, we adopted a similar methodology. We did use the minimal action set, and did add random ``no-ops'' to the beginning of episodes, but we opted not to utilize the life counter. 

\subsubsection{Computational Cost}

We found that Blob-PROST is far more computationally practical than DQN. We compared the algorithms' resource needs on 3.2GHz Intel Core i7-4790S CPUs. We ran them for 24 hours (to allow resource usage to stabilize), then measured their runtime and memory use.

The computational cost of our implementation of Blob-PROST can vary from game to game. The runtime of our implementation ranged from 56 decisions per second (\textsc{Alien}) to 300 decisions per second (\textsc{Star Gunner}), that is 280-1500 frames per second (many times real time speed). The memory utilization of our implementation ranged from 50MB (\textsc{Pong}) to 9GB (\textsc{Battle Zone}). Note that \textsc{Battle Zone} was an outlier; the next most memory intensive game (\textsc{Star Gunner}) used only 3.7GB of memory. Furthermore, the memory utilization of Blob-PROST can likely be effectively controlled by simplifying the color palette or through the use of feature hashing ({\em e.g.} \cite{Bellemare12b}).

In contrast, we found DQN's computational cost to be quite consistent across games, running at a speed of approximately 5 decisions per second ({\em i.e.} 20 frames per second, 3 times slower than real time), and requiring approximately 9.8GB of memory. DQN already uses a reduced color palette and is not immediately amenable to feature hashing to control memory usage. On the other hand, it is amenable to GPU acceleration to improve runtime. Mnih et al. do not report DQN's runtime but in recent follow-up work the GPU accelerated speed has been reported to be approximately 330 frames per second \cite{vanHasselt15}, still slower than the Blob-PROST agent in most games. Further note that obtaining enough GPUs to support multiple independent trials in all the games (necessary for statistical comparisons) is, in itself, a prohibitive cost. Emerging work~\cite{Mnih16} shows that a related approach can be accelerated via CPU parallelism.

\subsubsection{ALE Performance}

Because only one trial of DQN was reported in each game, and because of the prohibitively high computational cost of independently re-evaluating DQN, a principled comparison between our method and DQN is essentially impossible. Instead, we used a number of ad hoc measures aimed at forming an intuitive understanding of the relative capabilities of the two algorithms, based on available evidence.

First, for each game we record how many trials of Blob-PROST obtained evaluation scores greater than DQN's single trial. If we were to compare an algorithm to itself in this way, we would expect roughly 50\% of the trials to have greater performance. In Table \ref{table:dqnVblobprost} (available in the Appendix), the column marked ``\% trials $>$ DQN'' reports the results. The average percentage of trials better than DQN, across all games, was 41\%. That is, if you select a game uniformly randomly and run Blob-PROST, there is an estimated 41\% chance that it will surpass DQN's reported score. 

Second, in each game we compared Blob-PROST's middle trial (12th best) to the single reported DQN trial. If Blob-PROST's evaluation score compares favorably to DQN in this trial then this suggests that scores like DQN's are typical for Blob-PROST. Again, if this method were used to compare an algorithm to itself, one would expect the middle trial to be better in roughly half of the games. In Table \ref{table:dqnVblobprost} the column marked ``Middle trial'' reports the results. Blob-PROST's middle trial evaluation score was better than DQN's in 43\% (21/49) of the games. In 3 of the games in which it performed worse, Blob-PROST's score was not stastically significantly different than DQN's. So in 49\% of the games Blob-PROST's middle trial was either not different from or better than DQN's single reported trial.

Third, in each game we compared Blob-PROST's best trial to the single DQN trial. Since it is possible that some of DQN's trials are high-performing outliers, this is intended to give a sense of the level of performance Blob-PROST is capable of reaching, regardless of whether it can do so consistently. In 65\% (32/49) of the games, Blob-PROST's best trial's performance exceeded that of the DQN trial. Again, in three games in which Blob-PROST did worse, the difference was not statistically significant. So in 71\% of the games Blob-PROST's best trial was either not different from or better than DQN's single reported trial. 

Finally, DQN's main result was achieving at least 75\% of a human's performance in over half of the games (29/49). The best trial of Blob-PROST crossed this threshold in 34/49 games. The middle trial did so in only 20/49 games. 

The Blob-PROST representation was generated using simple and natural enhancements to BASS (one of DQN's main comparison points) aimed at incorporating similar structural biases to those encoded by DQN's designer-provided network architecture. All told, these results seem to indicate that this fixed representation is of comparable quality to DQN's learned representation across many games.

\section{ALE Benchmark} \label{benchmark}
As discussed above, besides better understanding the core issues underlying DQN's success in the ALE, it is important to present a reproducible benchmark without the problematic flaws discussed in Section~\ref{comparison_dqn} (\emph{i.e.} one that reports more than one trial, evaluates the final set of weights, and eschews game-specific prior knowledge). Moreover, establishing this benchmark is important for the ALE's continued use as an evaluation platform. The principled evaluation of complicated approaches depends upon the existence of a sound baseline for what can be achieved with simple approaches.

To this end we also present the performance of Blob-PROST using the full action set. Our agents' average performance after 24 trials using the full action set is reported in the last column of Table~4 in the Appendix. We recommend that future comparisons to Blob-PROST use these results. Surprisingly, we found that when using the full action set Blob-PROST performed slightly better in comparison to DQN using the measures described above than when using the minimal action set. It is not entirely clear why this would be; one possible explanation is that the presence of redundant actions may have a positive impact on exploration in some games. 

\section{Conclusions and Future Work}
While it is difficult to draw firm conclusions, the results indicate the Blob-PROST's performance in the ALE is competitive to DQN's, albeit likely slightly worse overall. Most importantly, these results indicate we may have been able to capture some of the key features of DQN in a practical, fixed linear representation. We saw progressive and dramatic improvements by respectively incorporating relative distances between objects, non-Markov features, and more sophisticated object detection. This illuminates some important representational issues that likely underly DQN's success. It also suggests that the general properties of the representations learned by DQN may be more important to its success in ALE than the specific features it learns in each game. This deeper understanding of the specific strengths of DQN should aid practitioners in determining when its benefits are likely to outweigh its costs.

It is important to note that we do not intend to diminish the importance of DQN as a seemingly stable combination of deep neural networks and RL. This is a promising and exciting research direction with potential applications in many domains. That said, taking into account the fact that DQN's performance may be inflated to an unknown degree as a result of evaluating the best performing set of weights, the fact of DQN's extreme computational cost, and the methodological issues underlying the reported DQN results, we conclude that Blob-PROST is a strong alternative candidate to DQN both as an ALE performance benchmark and as a platform on which to build future ALE agents when representation learning is not the topic being studied. Our results also suggest in general that fixed representations inspired by the principles underlying convolutional networks may yield competitive, lighter-weight alternatives.

As future work, it may be interesting to investigate the remaining discrepancies between our results and DQN's. In some games like \textsc{Gravitar, Frostbite} and \textsc{Krull} Blob-PROST's performance was several times higher than DQN's, while the opposite was true in games such as \textsc{Star Gunner, Breakout} and \textsc{Atlantis}. In general, DQN seems to excel in shooting games (a large part of the games supported by the ALE), maybe because it is able to easily encode object velocities and to predict objects' future positions. DQN also excels on games that require a more holistic view of the whole screen (e.g. \textsc{Breakout, Space Invaders}), something pairwise features struggle with. On the other hand, some of the games in which Blob-PROST surpasses DQN are games where it is fairly easy to die, maybe because DQN interrupts the learning process after the agent loses a life. Other games Blob-PROST succeeds in are games where the return is very sparse (e.g. \textsc{Tennis, Montezuma's Revenge}). These games may generate very small gradients for the network, making it harder for an algorithm to learn both the representation and a good policy. Alternatively DQN's process of sampling from the experience memory may not be effective to draw the ``interesting'' samples.

These conjectures suggest directions for future work. Adaptive representation methods may potentially benefit from these insights by building in stronger bias toward the types of features we have investigated here. This would allow them to quickly perform well, and then focus energy on learning exceptions to the rule. In the realm of linear methods, these results suggest that there may still be simple, generic enhancements that yield dramatic improvements. More sophisticated object detection may be beneficial, as might features that encode more holistic views of the screen.

\section*{Acknowledgements}
This research was supported by Alberta Innovates Technology Futures and the Alberta Innovates Centre for Machine Learning. Computing resources were provided by Compute Canada through CalculQu\'ebec.

\clearpage

\appendix
The following tables report the complete results discussed in the paper. Table~3 reports the data used to compare B-PROS results to other linear feature representations' reported results (see Section~3). Table~4 presents the results of the three feature sets introduced in this paper (B-PROS, B-PROST and Blob-PROST) as well as the results using Blob-PROST with the full action set, which we recommend to be used as benchmark in the future (see Sections~4,~5~and~7). Finally, Table~5 reports the data used to compare Blob-PROST to DQN's reported results (see Section \ref{comparison_dqn}).

\begin{table*}[h!]
\begin{center}
\caption{Control performance in Atari 2600 games when using linear function approximation. Bold indicates the best mean score in the first five columns, which were obtained using all the 18 actions available, after learning for 5,000 episodes and evaluating for 500 episodes~[2]. CAF denotes contingency awareness features~[3]. These results were reported after learning for 10,000 episodes and the average reported is the average score of the last 500 learning episodes. They are not directly comparable but note B-PROS outperforms CAF in the majority of games, even using half of the samples.}
\label{table:ale_lfa}
\small{
\begin{tabular}{|r|r|r|r|r|rl||r|}
\hline
Game                         & Basic           & BASS           & DISCO         & LSH             & \multicolumn{2}{c||}{B-PROS (std. dev.)}   & CAF          \\ \hline \hline
\textsc{Asterix}             & 862.3           & 859.8          & 754.6         & 987.3           & {\bf 4078.7}   & (1016.0)                 & 1332.0       \\ \hline
\textsc{Beam Rider}          & 929.4           & 872.7          & 563.0         & 793.6           & {\bf 1528.7}   & (504.4)                  & 1742.7       \\ \hline
\textsc{Breakout}            & 3.3             & 5.2            & 3.9           & 2.5             & {\bf 13.5}     & (11.0)                   & 6.1          \\ \hline
\textsc{Enduro}              & 111.8           & 129.1          & 0.0           & 95.8            & {\bf 240.5}    & (24.2)                   & 159.4        \\ \hline
\textsc{Freeway}             & 11.3            & 16.4           & 12.8          & 15.4            & {\bf 31.0}     & (0.7)                    & 20.0        \\ \hline
\textsc{Pong}                & -19.2           & -19.0          & -19.6         & -19.9           & {\bf 4.9 }     & (6.6)                    & -17.4        \\ \hline
\textsc{Q*Bert}              & 613.5           & 497.2          & 326.3         & 529.1           & {\bf 1099.8}   & (740.0)                  & 960.3        \\ \hline
\textsc{Seaquest}            & 579.0           & 664.8          & 421.9         & 508.5           & {\bf 1048.5}   & (321.2)                  & 675.5        \\ \hline
\textsc{Space Invaders}      & 203.6           & 250.1          & 239.1         & 222.2           & {\bf 384.7}    & (87.2)                   & 267.9        \\ \hline
\hline
\textsc{Alien}               & 939.2           & 893.4          & 623.6         & 510.2           & {\bf 1563.8}   & (529.6)                  & 103.2        \\ \hline
\textsc{Amidar}              & 64.9            & 103.4          & 67.9          & 45.1            & {\bf 174.7}    & (63.0)                   & 183.6        \\ \hline
\textsc{Assault}             & 465.8           & 378.4          & 371.7         & {\bf 628.0}     & 472.9          & (103.7)                  & 537.0          \\ \hline
\textsc{Asteroids}           & 829.7           & 800.3          & 744.5         & 590.7           & {\bf 1281.6}   & (182.5)                  & 89.0           \\ \hline
\textsc{Atlantis}            & {\bf 62687.0}   & 25375.0        & 20857.3       & 17593.9         & 33328.9        & (9787.2)                 & 852.9        \\ \hline
\textsc{Bank Heist}          & 98.8            & 71.1           & 51.4          & 64.6            & {\bf 214.5}    & (128.2)                  & 67.4         \\ \hline
\textsc{Battle Zone}         & 15534.3         & 12750.8        & 0.0           & 14548.1         & {\bf 18433.5}  & (4473.2)                 & 16.2         \\ \hline
\textsc{Berzerk}             & 329.2           & {\bf 491.3}    & 329.0         & 441.0           & 471.6          & (100.4)                  &  -           \\ \hline
\textsc{Bowling}             & 28.5            & {\bf 43.9}     & 35.2          & 26.1            & 36.9           & (11.7)                   & 36.4         \\ \hline
\textsc{Boxing}              & -2.8            & {\bf 15.5}     & 12.4          & 10.5            & 8.3            & (5.8)                    & 9.8          \\ \hline
\textsc{Centipede}           & 7725.5          & 8803.8         & 6210.6        & 6161.6          & {\bf 12258.6}  & (2064.0)                 & 4647.0         \\ \hline
\textsc{Chopper Command}     & 1191.4          & 1581.5         & 1349.0        & 943.0           & {\bf 2180.1}   & (584.5)                  & 16.9         \\ \hline
\textsc{Crazy Climber}       & 6303.1          & 7455.6         & 4552.9        & 20453.7         & {\bf 24669.0}  & (8526.0)                 & 149.8        \\ \hline
\textsc{Demon Attack}        & 520.5           & 318.5          & 208.8         & 355.8           & {\bf 664.2}    & (108.9)                  & 0.0          \\ \hline
\textsc{Double Dunk}         & -15.8           & -13.1          & -23.2         & -21.6           & {\bf -6.7}     & (2.5)                    & -16.0        \\ \hline
\textsc{Elevator Action}     & 3025.2          & 2377.6         & 4.6           & 3220.6          & {\bf 6879.5}   & (3610.5)                 & -            \\ \hline
\textsc{Fishing Derby}       & -92.6           & -92.1          & -89.5         & -93.2           & {\bf -87.8}    & (4.5)                    & -85.1        \\ \hline
\textsc{Frostbite}           & 161.0           & 161.1          & 176.6         & 216.9           & {\bf 768.8}    & (747.5)                  & 180.9        \\ \hline
\textsc{Gopher}              & 545.8           & 1288.3         & 295.7         & 941.8           & {\bf 5028.8}   & (1159.3)                 & 2368         \\ \hline
\textsc{Gravitar}            & 185.3           & 251.1          & 197.4         & 105.9           & {\bf 613.8}    & (142.3)                  & 429          \\ \hline
\textsc{H.E.R.O.}            & 6053.1          & 6458.8         & 2719.8        & 3835.8          & {\bf 8086.0}   & (2034.2)                 & 7295         \\ \hline
\textsc{Ice Hockey}          & -13.9           & -14.8          & -18.9         & -15.1           & {\bf 0.6}      & (1.9)                    & -3.2         \\ \hline
\textsc{James Bond}          & 197.3           & 202.8          & 17.3          & 77.1            & {\bf 336.6}    & (107.3)                  & 354.1        \\ \hline
\textsc{Journey Escape}      & -8441.0         & -14730.7       & -9392.2       & -13898.9        & {\bf -3953.9}  & (2968.2)                 & -            \\ \hline
\textsc{Kangaroo}            & 962.4           & {\bf 1622.1}   & 457.9         & 256.4           & 1337.6         & (628.2)                  & 8.8          \\ \hline
\textsc{Krull}               & 2823.3          & 3371.5         & 2350.9        & 2798.1          & {\bf 4446.6}   & (999.3)                  & 3341.0       \\ \hline
\textsc{Kung-Fu Master}      & 16416.2         & 19544.0        & 3207.0        & 8715.6          & {\bf 25255.8}  & (3848.9)                 & 29151.0      \\ \hline
\textsc{Montezuma's Revenge} & {\bf 10.7}      & 0.1            & 0.0           & 0.1             & 0.1            & (0.3)                    & 259.0        \\ \hline
\textsc{Ms. Pac-Man}         & 1537.2          & 1691.8         & 999.6         & 1070.8          & {\bf 3494.1}   & (471.3)                  & 1227.0       \\ \hline
\textsc{Name This Game}      & 1818.9          & 2386.8         & 1951.0        & 2029.8          & {\bf 6857.4}   & (477.1)                  & 2247.0       \\ \hline
\textsc{Pooyan}              & 800.3           & 1018.9         & 402.7         & 1225.3          & {\bf 1461.7}   & (192.3)                  & -            \\ \hline
\textsc{Private Eye}         & 81.9            & 100.7          & -23.0         & {\bf 684.3}     & 148.8          & (151.1)                  & 86.0         \\ \hline
\textsc{River Raid}          & 1708.9          & 1438.0         & 0.0           & 1904.3          & {\bf 4228.0}   & (697.8)                  & 2650.0       \\ \hline
\textsc{Road Runner}         & 67.7            & 65.2           & 21.4          & 42.0            & {\bf 17891.9}  & (3795.6)                 & 89.1         \\ \hline
\textsc{Robotank}            & 12.8            & 10.1           & 9.3           & 10.8            & {\bf 18.5}     & (1.5)                    & 12.4         \\ \hline
\textsc{Star Gunner}         & 850.2           & {\bf 1069.5}   & 1002.2        & 722.9           & 1040.5         & (63.1)                   & 9.4          \\ \hline
\textsc{Tennis}              & -0.2            & -0.1           & -0.1          & -0.1            & {\bf 0.0}      & (0.0)                    & 0.0          \\ \hline
\textsc{Time Pilot}          & 1728.2          & 2299.5         & 0.0           & {\bf 2429.2}    & 1813.8         & (696.4)                  & 24.9         \\ \hline
\textsc{Tutankham}           & 40.7            & 52.6           & 0.0           & 85.2            & {\bf 119.0}    & (31.7)                   & 98.2         \\ \hline
\textsc{Up 'n Down}         & 3532.7          & 3351.0         & 2473.4        & 2475.1          & {\bf 7737.7}   & (1936.2)                 & 2449.0       \\ \hline
\textsc{Venture}             & 0.0             & {\bf 66.0}     & 0.0           & 0.0             & 0.0            & (0.0)                    & 0.6          \\ \hline
\textsc{Video Pinball}       & 15046.8         & 12574.2        & 10779.5       & 9813.9          & {\bf 18086.3}  & (5195.4)                 & 19761.0      \\ \hline
\textsc{Wizard of Wor}       & 1768.8          & {\bf 1981.3}   & 935.6         & 945.5           & 1900.6         & (619.4)                  & 36.9         \\ \hline
\textsc{Zaxxon}              & 1392.0          & 2069.1         & 69.8          & 3365.1          & {\bf 4465.1}   & (2856.4)                 & 21.4         \\ \hline \hline
Times Best                   & 2               & 7              & 0             & 3               & \multicolumn{2}{c||}{41}                                 \\
\cline{1-7}
\end{tabular}
}
\end{center}
\end{table*}

\begin{table*}[h!]
\begin{center}
\label{table:ale_features_comparison}
\caption{Comparison between the three proposed feature sets. See Section~\ref{empirical_lfa} for details. The last column presents the average performance of Blob-PROST using the full action set for comparison to future work (see Section \ref{benchmark}). The standard deviation reported represents the variability over the 24 independent learning trials. The highest average among the first three columns is in bold face. We use $^\dagger$ to denote statistically different averages between B-PROST and Blob-PROST, which were obtained with Welch's t-tests (p$<$0.05).}
\small{
\begin{tabular}{|r|rl|rl|rl|rl|}
\hline
\multirow{2}{*}{Game}         &\multicolumn{2}{c|}{{B-PROS}}             &\multicolumn{2}{c|}{B-PROST}                 &\multicolumn{2}{c|}{Blob-PROST (min.)}  &\multicolumn{2}{c|}{Blob-PROST (full)}\\
                              &\multicolumn{2}{c|}{{Avg. (std. dev.)}}   &\multicolumn{2}{c|}{{Avg. (std. dev.)}}      &\multicolumn{2}{c|}{{Avg. (std. dev.)}} &\multicolumn{2}{c|}{{Avg. (std. dev.)}} \\ \hline \hline

\textsc{Asterix}              &8194.3         &(1802.9)    &{\bf 8928.8}$^\dagger$   &(1714.5)  &4400.8                     &(1201.3)  &3996.6         &(743.9)  \\ \hline
\textsc{Beam Rider}           &1686.9         &(255.2)     &1808.9                   &(309.2)   &{\bf 1902.3}               &(423.9)   &2367.3         &(815.0)  \\ \hline
\textsc{Breakout}             &7.1            &(1.7)       &15.0                     &(4.7)     &{\bf 46.7}$^\dagger$       &(45.7)    &52.9           &(38.0)   \\ \hline
\textsc{Enduro}               &34.9           &(71.5)      &207.7                    &(23.1)    &{\bf 257.0}$^\dagger$      &(79.8)    &296.7          &(7.8)    \\ \hline
\textsc{Freeway}              &23.8           &(6.7)       &29.1                     &(6.3)     &{\bf 31.5}$^\dagger$       &(1.2)     &32.3           &(0.5)    \\ \hline
\textsc{Pong}                 &10.9           &(5.2)       &18.9                     &(1.3)     &{\bf 20.1}$^\dagger$       &(0.5)     &20.2           &(0.4)    \\ \hline
\textsc{Q*Bert}               &3647.8         &(1273.3)    &3608.7                   &(1129.0)  &{\bf 6946.8}$^\dagger$     &(3036.4)  &8072.4         &(2210.5) \\ \hline
\textsc{Seaquest}             &1366.1         &(445.9)     &1636.5                   &(519.5)   &{\bf 1664.2}               &(440.4)   &1664.2         &(440.4)  \\ \hline
\textsc{Space Invaders}       &505.1          &(130.4)     &582.9                    &(139.0)   &{\bf 723.1}$^\dagger$      &(101.5)   &844.8          &(144.9)  \\ \hline
\hline
\textsc{Alien}                &3169.5         &(616.8)     &4074.5                   &(533.7)   &{\bf 4154.8}               &(532.5)   &4154.8         &(532.5)  \\ \hline
\textsc{Amidar}               &337.9          &(77.8)      &308.2                    &(121.3)   &{\bf 486.7}$^\dagger$      &(195.4)   &408.4          &(177.5)  \\ \hline
\textsc{Assault}              &1329.4         &(215.3)     &{\bf 1485.9}$^\dagger$   &(416.9)   &1356.0                     &(299.6)   &1107.9         &(207.7)  \\ \hline
\textsc{Asteroids}            &1145.7         &(139.9)     &1306.7                   &(198.6)   &{\bf 1673.5}$^\dagger$     &(211.1)   &1759.5         &(182.1)  \\ \hline
\textsc{Atlantis}             &\bf{33465.9}   &(13879.0)   &19195.9                  &(15223.3) &19347.2                    &(10478.5) &37428.5        &(11599.7)\\ \hline
\textsc{Bank Heist}           &227.7          &(127.8)     &332.5                    &(53.1)    &{\bf 463.4}$^\dagger$      &(93.6)    &463.4          &(93.6)   \\ \hline
\textsc{Battle Zone}          &22794.8        &(4643.4)    &25321.8                  &(5063.5)  &\bf{26222.8}               &(4070.0)  &26222.8        &(4070.0) \\ \hline
\textsc{Bowling}              &36.5           &(10.6)      &48.0                     &(20.0)    &{\bf 59.1}$^\dagger$       &(18.6)    &65.9           &(14.2)   \\ \hline
\textsc{Boxing}               &27.2           &(30.1)      &26.1                     &(25.6)    &{\bf 89.4}$^\dagger$       &(16.5)    &89.4           &(16.5)   \\ \hline
\textsc{Carnival}             &-              &-           &-                        &-         &-                          &-         &4322.0         &(3705.0) \\ \hline
\textsc{Centipede}            &{\bf 18871.4}  &(2624.9)    &2579.9                   &(4848.7)  &3903.3                     &(6838.8)  &3903.3         &(6838.8) \\ \hline
\textsc{Chopper Command}      &3884.9         &(872.5)     &{\bf 4072.3}$^\dagger$   &(931.0)   &3006.6                     &(782.0)   &3006.6         &(782.0)  \\ \hline
\textsc{Crazy Climber}        &24763.4        &(8080.5)    &34486.7                  &(13056.1) &{\bf 59514.3}$^\dagger$    &(13721.0) &73241.5        &(10467.9)\\ \hline
\textsc{Demon Attack}         &2085.4         &(532.5)     &{\bf 2441.9}$^\dagger$   &(571.9)   &1745.8                     &(256.0)   &1441.8         &(184.8)  \\ \hline
\textsc{Double Dunk}          &-5.5           &(3.0)       &{\bf 5.9}$^\dagger$      &(13.3)    &-6.4                       &(0.9)     &-6.4           &(0.9)    \\ \hline
\textsc{Fishing Derby}        &-86.5          &(5.5)       &-82.8                    &(6.2)     &{\bf -58.8}$^\dagger$      &(12.0)    &-58.8          &(12.0)   \\ \hline
\textsc{Frostbite}            &1974.9         &(1046.3)    &2382.9                   &(1182.8)  &{\bf 3389.7}$^\dagger$     &(743.6)   &3389.7         &(743.6)  \\ \hline
\textsc{Gopher}               &3784.7         &(670.7)     &{\bf 5272.4}             &(1077.0)  &5076.6                     &(1235.1)  &6823.4         &(1022.5) \\ \hline
\textsc{Gravitar}             &778.2          &(191.8)     &788.7                    &(350.7)   &{\bf 1231.8}$^\dagger$     &(423.4)   &1231.8         &(423.4)  \\ \hline
\textsc{H.E.R.O.}             &9572.1         &(4160.2)    &13552.5                  &(1867.4)  &{\bf 13690.3}              &(4291.2)  &13690.3        &(4291.2) \\ \hline
\textsc{Ice Hockey}           &2.1            &(1.7)       &7.3                      &(2.3)     &{\bf 14.5}$^\dagger$       &(3.4)     &14.5           &(3.4)    \\ \hline
\textsc{James Bond}           &378.0          &(128.9)     &373.5                    &(158.3)   &{\bf 636.3}$^\dagger$      &(191.8)   &636.3          &(191.8)  \\ \hline
\textsc{Kangaroo}             &4946.0         &(2694.3)    &{\bf 5898.4}$^\dagger$   &(3166.3)  &3800.3                     &(2211.0)  &3800.3         &(2211.0) \\ \hline
\textsc{Krull}                &5247.1         &(957.4)     &5973.8                   &(904.6)   &{\bf 8333.9}$^\dagger$     &(5599.5)  &8333.9         &(5599.5) \\ \hline
\textsc{Kung-Fu Master}       &34025.7        &(5994.3)    &{\bf 35173.0}            &(6190.8)  &34075.6                    &(7716.7)  &33868.5        &(6247.5) \\ \hline
\textsc{Montezuma's Revenge}  &300.3          &(155.7)     &345.8                    &(433.6)   &{\bf 778.1}$^\dagger$      &(789.8)   &778.1          &(789.8)  \\ \hline
\textsc{Ms. Pac-Man}          &4402.9         &(905.4)     &{\bf 5544.1}$^\dagger$   &(727.9)   &4625.6                     &(774.3)   &4625.6         &(774.3)  \\ \hline
\textsc{Name This Game}       &5697.7         &(809.7)     &{\bf 6593.1}$^\dagger$   &(750.9)   &5883.7                     &(1542.1)  &6580.1         &(1773.0) \\ \hline
\textsc{Pooyan}               &-              &-           &-                        &-         &-                          &-         &2228.1         &(274.5)  \\ \hline
\textsc{Private Eye}          &106.3          &(42.9)      &{\bf 502.7}$^\dagger$    &(1842.1)  &33.0                       &(47.6)    &33.0           &(47.6)   \\ \hline
\textsc{River Raid}           &9171.4         &(973.6)     &9928.8                  &(1,443.6)  &{\bf 10629.1}              &(2110.5)  &10629.1        &(2110.5) \\ \hline
\textsc{Road Runner}          &25342.7        &(4168.4)    &{\bf 32282.9}$^\dagger$  &(5707.8)  &24558.3                    &(12369.2) &24558.3        &(12369.2)\\ \hline
\textsc{Robotank}             &21.4           &(1.3)       &22.6                     &(2.4)     &{\bf 28.3}$^\dagger$       &(2.7)     &28.3           &(2.7)    \\ \hline
\textsc{Skiing}               &-              &-           &-                        &-         &-                          &-         &-29842.6       &(19.8)   \\ \hline
\textsc{Star Gunner}          &{\bf 1355.3}   &(343.8)     &1020.1                   &(304.8)   &1227.7$^\dagger$           &(165.1)   &1227.7         &(165.1)  \\ \hline
\textsc{Tennis}               &-0.2           &(0.8)       &{\bf 0.0}                &(0.0)     &{\bf 0.0}                  &(0.0)     &0.0            &(0.0)    \\ \hline
\textsc{Time Pilot}           &2501.2         &(943.4)     &2154.6                   &(875.4)   &{\bf 3933.1}$^\dagger$     &(713.9)   &3972.0         &(878.3)  \\ \hline
\textsc{Tutankham}            &{\bf 110.3}    &(40.8)      &90.0$^\dagger$           &(52.4)    &61.2                       &(61.3)    &81.4           &(50.5)   \\ \hline
\textsc{Up'n Down}            &6140.4         &(5418.7)    &7888.2                   &(5950.1)  &{\bf 12197.0}              &(14027.3) &19533.0        &(18733.6)\\ \hline
\textsc{Venture}              &97.3           &(223.7)     &182.1                    &(370.1)   &{\bf 244.5}                &(490.4)   &244.5          &(490.4)  \\ \hline
\textsc{Video Pinball}        &14239.4        &(4314.0)    &{\bf 19358.3}$^\dagger$  &(3006.2)  &10006.3                    &(5860.8)  &9783.9         &(3043.9) \\ \hline
\textsc{Wizard of Wor}        &2500.8         &(754.7)     &2893.6                   &(842.6)   &{\bf 3174.8}               &(1227.9)  &2733.6         &(1259.0) \\ \hline
\textsc{Zaxxon}               &10993.8        &(2118.2)    &{\bf 11599.8}$^\dagger$  &(2660.7)  &8204.4                     &(2845.3)  &8204.4         &(2845.3) \\ \hline \hline
\textsc{Times best}           &\multicolumn{2}{c|}{4}      &\multicolumn{2}{c|}{14.5}  &\multicolumn{2}{c|}{30.5}        \\ \cline{1-7}
\textsc{Times statistically best} &\multicolumn{2}{c|}{N/A}      &\multicolumn{2}{c|}{14}  &\multicolumn{2}{c|}{23}      \\ \cline{1-7}

\end{tabular}
}
\end{center}
\end{table*}

\begin{table*}[h!]
\begin{center}
\caption{The first column presents the average performance of Blob-PROST using the minimal action set. The standard deviation reported represents the variability over the 24 independent learning trials. The rest of the columns present our comparison between Blob-PROST and DQN (see Section~\ref{comparison_dqn} for details). In these results the Blob-PROST agent uses the minimal action set, for the sake of comparison. The standard deviation reported for the best trial as well as for the median trial is the standard deviation while evaluating for 499 episodes. We use bold face to denote values that are higher than DQN's and $^\dagger$ to denote Blob-PROST averages that are statistically significantly different than DQN's average. To do the comparison for each game we used Welch's t-test (p$<$0.025, which accounts for the usage of DQN results in two tests, for an overall significance level of 0.05).}
\label{table:dqnVblobprost}
\tiny{
\begin{tabular}{|r|rl|rl|rl|r|rl|r|r|}
\hline
\multirow{2}{*}{Game}         &\multicolumn{2}{c|}{\multirow{2}{*}{Avg. (std. dev.)}}     &\multicolumn{2}{c|}{\multirow{2}{*}{Best trial (std. dev.)}}  &\multicolumn{2}{c|}{\multirow{2}{*}{Middle trial (std. dev.)}}  &\% trials   &\multicolumn{2}{c|}{\multirow{2}{*}{DQN}}  &\multirow{2}{*}{Human}  &\multirow{2}{*}{Random} \\
                              &               &           &                            &           &                            &               & $>$ DQN   &         &            &           &           \\ \hline \hline
\textsc{Asterix}              &4400.8         &(1201.3)    &{\bf 6921.5}                &(4501.2)   &4472.9$^\dagger$            &(2514.8)       &8.3        &6012.0   &(1744.0)    &8503.0     &210.0      \\ \hline
\textsc{Beam Rider}           &1902.3         &(423.9)    &2965.5$^\dagger$            &(1376.1)   &1807.6$^\dagger$            &(657.0)        &0.0        &6846.0   &(1619.0)    &5775.0     &363.9      \\ \hline
\textsc{Breakout}             &46.7           &(45.7)     &190.3$^\dagger$             &(179.7)    &33.2$^\dagger$              &(21.3)         &0.0        &401.2    &(26.9)      &31.8       &1.7        \\ \hline
\textsc{Enduro}               &257.0          &(79.8)      &299.1                       &(26.9)     &279.2$^\dagger$             &(28.2)         &0.0        &301.8    &(24.6)      &309.6      &0.0        \\ \hline
\textsc{Freeway}              &31.5           &(1.2)      &{\bf 32.6}$^\dagger$        &(1.0)      &{\bf 31.7}$^\dagger$        &(1.0)          &95.8       &30.3     &(0.7)       &29.6       &0.0        \\ \hline
\textsc{Pong}                 &20.1           &(0.5)      &{\bf 20.5}$^\dagger$        &(0.8)      &{\bf 20.2}$^\dagger$        &(1.4)          &95.8       &18.9     &(1.3)       &9.3        &-20.7      \\ \hline
\textsc{Q*Bert}               &6946.8         &(3036.4)   &{\bf 14449.4}$^\dagger$     &(4111.7)   &5881.4$^\dagger$            &(1069.0)       &12.5       &10596.0  &(3294.0)    &13455.0    &163.9      \\ \hline
\textsc{Seaquest}             &1664.2         &(440.4)    &2278.0$^\dagger$            &(294.1)    &1845.1$^\dagger$            &(469.8)        &0.0        &5286.0   &(1310.0)    &20182.0    &68.4       \\ \hline
\textsc{Space Invaders}       &723.1          &(101.5)    &889.8$^\dagger$             &(374.5)    &712.9$^\dagger$             &(250.3)        &0.0        &1976.0   &(893.0)     &1652.0     &148.0      \\ \hline
\hline
\textsc{Alien}                &4154.8         &(532.5)    &{\bf 4886.6}$^\dagger$      &(1151.9)   &{\bf 4219.1}$^\dagger$      &(635.4)        &95.8       &3069.0   &(1093.0)    &6875.0     &227.8      \\ \hline
\textsc{Amidar}               &486.7          &(195.4)    &{\bf 825.4}                 &(225.3)    &522.3                       &(153.1)        &12.5       &739.5    &(3024.0)    &1676.0     &5.8        \\ \hline
\textsc{Assault}              &1356.0         &(299.6)    &1829.3$^\dagger$            &(702.9)    &1380.8$^\dagger$            &(449.9)        &0.0        &3359.0   &(775.0)     &1496.0     &224.4      \\ \hline
\textsc{Asteroids}            &1673.5         &(211.1)    &{\bf 2229.9}$^\dagger$      &(937.0)    &{\bf 1635.1}                &(617.9)        &50.0       &1629.0   &(542.0)     &13157.0    &719.1      \\ \hline
\textsc{Atlantis}             &19347.2        &(10478.5)  &42937.7$^\dagger$           &(13763.1)  &19983.2$^\dagger$           &(5830.4)       &0.0        &85641.0  &(17600.0)   &29028.0    &12850.0    \\ \hline
\textsc{Bank Heist}           &463.4          &(93.6)     &{\bf 793.6}$^\dagger$       &(102.9)    &{\bf 455.1}                 &(100.9)        &62.5       &429.7    &(650.0)     &734.4      &14.2       \\ \hline
\textsc{Battle Zone}          &26222.8        &(4070.0)   &{\bf 37850.0}$^\dagger$     &(7445.0)   &25836.0                    &(6616.3)       &41.7       &26300.0  &(7725.0)    &37800.0    &2360.0     \\ \hline
\textsc{Bowling}              &59.1           &(18.6)     &{\bf 91.1}$^\dagger$        &(13.6)     &{\bf 62.0}                  &(2.5)          &91.7       &42.4     &(88.0)      &154.8      &23.1       \\ \hline
\textsc{Boxing}               &89.4           &(16.5)     &{\bf 98.3}$^\dagger$        &(3.3)      &{\bf 95.8}$^\dagger$        &(5.5)          &87.5       &71.8     &(8.4)       &4.3        &0.1        \\ \hline
\textsc{Centipede}            &3903.3         &(6838.8)   &{\bf 21137.0}$^\dagger$     &(6722.8)   &426.7$^\dagger$             &(313.8)        &20.8       &8309.0   &(5237.0)    &11963.0    &2091.0     \\ \hline
\textsc{Chopper Command}      &2998.0         &(763.1)    &4898.9$^\dagger$            &(2511.2)   &2998.0$^\dagger$            &(1197.7)       &0.0        &6687.0   &(2916.0)    &9882.0     &811.0      \\ \hline
\textsc{Crazy Climber}        &59514.3        &(13721.0)  &81016.0$^\dagger$           &(26529.2)  &59848.9$^\dagger$           &(27960.7)      &0.0        &114103.0 &(22797.0)   &35411.0    &10781.0    \\ \hline
\textsc{Demon Attack}         &1745.8         &(256.0)    &2166.0$^\dagger$            &(1327.5)   &1774.2$^\dagger$            &(979.8)        &0.0        &9711.0   &(2406.0)    &3401.0     &152.1      \\ \hline
\textsc{Double Dunk}          &-6.4           &(0.9)      &{\bf -4.1}$^\dagger$        &(2.3)      &{\bf -6.2}$^\dagger$        &(2.8)          &100.0      &-18.1    &(2.6)       &-15.5      &-18.6      \\ \hline
\textsc{Fishing Derby}        &-58.8          &(12.0)     &-28.7$^\dagger$             &(19.0)     &-57.4$^\dagger$             &(18.3)         &0.0        &-0.8     &(19.0)      &5.5        &-91.7      \\ \hline
\textsc{Frostbite}            &3389.7         &(743.6)    &{\bf 4534.0}$^\dagger$      &(1496.8)   &{\bf 3557.5}$^\dagger$      &(864.3)        &100.0      &328.3    &(250.5)     &4335.0     &65.2       \\ \hline
\textsc{Gopher}               &5076.6         &(1235.1)   &7451.1$^\dagger$            &(3146.1)   &5006.9$^\dagger$            &(3118.9)       &0.0        &8520.0   &(32.8)      &2321.0     &257.6      \\ \hline
\textsc{Gravitar}             &1231.8         &(423.4)    &{\bf 1709.7}$^\dagger$      &(669.7)    &{\bf 1390.6}$^\dagger$      &(752.9)        &91.7       &306.7    &(223.9)     &2672.0     &173.0      \\ \hline
\textsc{H.E.R.O.}             &13690.3        &(4291.2)   &{\bf 20273.1}$^\dagger$     &(1155.1)   &13642.1$^\dagger$           &(48.4)         &12.5       &19950.0  &(158.0)     &25673.0    &1027.0     \\ \hline
\textsc{Ice Hockey}           &14.5           &(3.4)      &{\bf 22.8}$^\dagger$        &(6.5)      &{\bf 14.5}$^\dagger$        &(5.5)          &100.0      &-1.6     &(2.5)       &0.9        &-11.2      \\ \hline
\textsc{James Bond}           &636.3          &(191.8)    &{\bf 1030.5}$^\dagger$      &(947.1)    &{\bf 587.8}                 &(99.5)         &50.0       &576.7    &(175.5)     &406.7      &29.0       \\ \hline
\textsc{Kangaroo}             &3800.3         &(2211.0)   &{\bf 9492.8}$^\dagger$      &(2918.2)   &3839.1$^\dagger$            &(1601.1)       &8.3        &6740.0   &(2959.0)    &3035.0     &52.0       \\ \hline
\textsc{Krull}                &8333.9         &(5599.5)   &{\bf 33263.4}$^\dagger$     &(15403.3)  &{\bf 7463.9}$^\dagger$      &(1719.7)       &95.8       &3805.0   &(1033.0)    &2395.0     &1598.0     \\ \hline
\textsc{Kung-Fu Master}       &34075.6        &(7716.7)   &{\bf 51007.6}$^\dagger$     &(9131.9)   &{\bf 33232.1}$^\dagger$     &(7904.5)       &91.7       &23270.0  &(5955.0)    &22736.0    &258.5      \\ \hline
\textsc{Montezuma's Revenge}  &778.1          &(789.8)    &{\bf 2508.4}$^\dagger$      &(27.8)     &{\bf 400.0}$^\dagger$       &(0.0)          &100.0      &0.0      &(0.0)       &4367.0     &0.0        \\ \hline
\textsc{Ms. Pac-Man}          &4625.6         &(774.3)    &{\bf 5917.9}$^\dagger$      &(1984.0)   &{\bf 4667.4}$^\dagger$      &(2126.0)       &100.0      &2311.0   &(525.0)     &15693.0    &307.3      \\ \hline
\textsc{Name This Game}       &5883.7         &(1542.1)   &{\bf 7787.0}$^\dagger$      &(744.7)    &6387.1$^\dagger$            &(1046.7)       &20.8       &7257.0   &(547.0)     &4076.0     &2292.0     \\ \hline
\textsc{Private Eye}          &33.0           &(47.6)     &100.3                        &(4.0)      &0.0                         &(0.0)          &0.0        &1788.0   &(5473.0)    &69571.0    &24.9       \\ \hline
\textsc{River Raid}           &10629.1        &(2110.5)   &{\bf 14583.3}$^\dagger$     &(3078.3)   &{\bf 9540.4}$^\dagger$      &(1053.8)       &91.7       &8316.0   &(1049.0)    &13513.0    &1339.0     \\ \hline
\textsc{Road Runner}          &24572.3        &(12381.1)  &{\bf 41828.0}$^\dagger$     &(9408.1)   &{\bf 28848.5}$^\dagger$     &(6857.4)       &75.0       &18257.0  &(4268.0)    &7845.0     &11.5       \\ \hline
\textsc{Robotank}             &28.3           &(2.7)      &34.4$^\dagger$              &(7.7)      &28.5$^\dagger$              &(8.0)          &0.0        &51.6     &(4.7)       &11.9       &2.2        \\ \hline
\textsc{Star Gunner}          &1227.7         &(165.1)    &1651.6$^\dagger$            &(435.4)    &1180.0$^\dagger$            &(85.2)         &0.0        &57997.0  &(3152.0)    &10250.0    &664.0      \\ \hline
\textsc{Tennis}               &0.0            &(0.0)      &{\bf 0.0}$^\dagger$         &(0.2)      &{\bf 0.0}$^\dagger$         &(0.2)          &100.0      &-2.5     &(1.9)       &-8.9       &-23.8      \\ \hline
\textsc{Time Pilot}           &3933.1         &(713.9)    &5429.5                      &(1659.7)   &4069.3$^\dagger$            &(647.7)        &0.0        &5947.0   &(1600.0)    &5925.0     &3568.0     \\ \hline
\textsc{Tutankham}            &61.2           &(61.3)     &{\bf 217.7}$^\dagger$       &(33.9)     &41.7$^\dagger$              &(4.2)          &8.3        &186.7    &(41.9)      &167.7      &11.4       \\ \hline
\textsc{Up and Down}          &12197.0        &(14027.3)  &{\bf 41257.8}$^\dagger$     &(12061.8)  &3716.7$^\dagger$            &(1050.8)       &45.8       &8456.0   &(3162.0)    &9082.0     &533.4      \\ \hline
\textsc{Venture}              &244.5          &(490.4)    &{\bf 1397.0}$^\dagger$      &(294.4)    &0.0$^\dagger$               &(0.0)          &20.8       &380.0    &(238.6)     &1188.0     &0.0        \\ \hline
\textsc{Video Pinball}        &10006.3         &(5860.8)   &21313.0$^\dagger$           &(14916.4)  &9480.7$^\dagger$            &(6563.1)       &0.0        &42684.0  &(16287.0)   &17298.0    &16257.0    \\ \hline
\textsc{Wizard of Wor}        &3174.8         &(1227.9)   &{\bf 5681.2}$^\dagger$      &(4573.2)   &{\bf 3442.9}                &(2582.0)       &50.0       &3393.0   &(2019.0)    &4757.0     &563.5      \\ \hline
\textsc{Zaxxon}               &8204.4         &(2845.3)   &{\bf 11721.8}$^\dagger$     &(4009.7)   &{\bf 9280.0}$^\dagger$      &(2251.4)       &87.5       &4977.0   &(1235.0)    &9173.0     &32.5       \\ \hline
\textsc{Times $>$ DQN}        &\multicolumn{2}{c|}{N/A} &\multicolumn{2}{c|}{32/49}    &\multicolumn{2}{c|}{21/49} \\ \cline{1-7}
\textsc{Times statist. $\geq$ DQN} &\multicolumn{2}{c|}{N/A} &\multicolumn{2}{c|}{35/49}    &\multicolumn{2}{c|}{24/49} \\ \cline{1-7}

\end{tabular}
}
\end{center}
\end{table*}

\clearpage
%
%
\clearpage


\end{document}